\documentclass[sigconf]{acmart}

\usepackage{amsmath}
\usepackage{multirow}
\usepackage{pseudocode}
\usepackage{subfigure}
\usepackage{multirow}
\usepackage{mdwlist}
\usepackage{booktabs}
\usepackage{latexsym}
\usepackage{tabularx}
\usepackage{fixltx2e}
\usepackage{microtype}
\usepackage{arydshln}

\usepackage[T1]{fontenc}
\usepackage[utf8]{inputenc}

\usepackage{textcomp}

\makeatletter
\def\adl@drawiv#1#2#3{%
        \hskip.5\tabcolsep
        \xleaders#3{#2.5\@tempdimb #1{1}#2.5\@tempdimb}%
                #2\z@ plus1fil minus1fil\relax
        \hskip.5\tabcolsep}
\newcommand{\cdashlinelr}[1]{%
  \noalign{\vskip\aboverulesep
           \global\let\@dashdrawstore\adl@draw
           \global\let\adl@draw\adl@drawiv}
  \cdashline{#1}
  \noalign{\global\let\adl@draw\@dashdrawstore
           \vskip\belowrulesep}}
\makeatother

\setcopyright{rightsretained}

\acmConference[SIGIR'18]{SIGIR}{July 2018}{Ann Arbor, Michigan, USA}
\acmYear{2018}
\copyrightyear{2018}

\acmArticle{x}
\acmPrice{15.00}

\begin{document}
\title[Unsupervised Cross-Lingual Information Retrieval]{Unsupervised Cross-Lingual Information Retrieval \\ using Monolingual Data Only}




\author{Robert Litschko}
\affiliation{\institution{University of Mannheim}}
\email{litschko@informatik.uni-mannheim.de}

\author{Goran Glava\v{s}}
\affiliation{\institution{University of Mannheim}}
\email{goran@informatik.uni-mannheim.de}

\author{Simone Paolo Ponzetto}
\affiliation{\institution{University of Mannheim}}
\email{simone@informatik.uni-mannheim.de}

\author{Ivan Vuli\'c}
\affiliation{\institution{University of Cambridge}}
\email{iv250@cam.ac.uk}

\renewcommand{\shortauthors}{R. Litschko et al.}

\begin{abstract}
We propose a fully unsupervised framework for ad-hoc cross-lingual information retrieval (CLIR) which requires no bilingual data at all. The framework leverages shared cross-lingual word embedding spaces in which terms, queries, and documents can be represented, irrespective of their actual language. The shared embedding spaces are induced solely on the basis of monolingual corpora in two languages through an iterative process based on adversarial neural networks. Our experiments on the standard CLEF CLIR collections for three language pairs of varying degrees of language similarity (English-Dutch/Italian/Finnish) demonstrate the usefulness of the proposed fully unsupervised approach. Our CLIR models with unsupervised cross-lingual embeddings outperform baselines that utilize cross-lingual embeddings induced relying on word-level and document-level alignments. We then demonstrate that further improvements can be achieved by unsupervised ensemble CLIR models. We believe that the proposed framework is the first step towards development of effective CLIR models for language pairs and domains where parallel data are scarce or non-existent.
\end{abstract}

%
%



\maketitle

\section{Introduction}

Retrieving relevant content across languages (i.e., \textit{cross-lingual information retrieval}, termed CLIR henceforth) requires the ability to bridge the lexical gap between languages \cite{Levow:2005ipm,Nie:2010book}. Traditional IR methods based on sparse text representations are not suitable for CLIR, since languages, in general, do not share much of the vocabulary. Even in the monolingual IR, they cannot bridge the lexical gap, being incapable of semantic generalization \cite{Landauer:19971}. A solution is to resort to structured real-valued semantic representations, that is, \textit{text embeddings} \cite{Landauer:19971,Mikolov:2013nips,Bojanowski:2017tacl}: these representations allow to generalize over the vocabularies observed in labelled data, and hence offer additional retrieval evidence and mitigate the ubiquitous problem of data sparsity. Their usefulness has been proven for monolingual \cite{Mitra:2017arxiv} and cross-lingual ad-hoc IR models \cite{Vulic:2015sigir}.

Besides the embedding-based CLIR paradigms, other approaches to bridging the lexical gap for CLIR exist. \textbf{1)} Full-blown Machine Translation (MT) systems are employed to translate either queries or documents \cite{Levow:2005ipm,Martino:2017sigir}, but these require huge amounts of parallel data, while such resources are still scarce for many language pairs and domains. \textbf{2)} The lexical chasm can be crossed by grounding queries and documents in an external multilingual knowledge source (e.g., Wikipedia or BabelNet) \cite{sorg2012exploiting,franco2014knowledge}. However, the concept coverage is limited for resource-lean languages, and all content not present in a knowledge base is effectively ignored by a CLIR system. 

Bilingual text embeddings, while displaying a wider applicability and versatility than the two other paradigms, still suffer from one important limitation: a \textit{bilingual supervision signal} is required to induce \textit{shared cross-lingual semantic spaces}. This supervision takes form of sentence-aligned parallel data \cite{hermann-blunsom:2014:P14-1}, pre-built word translation pairs \cite{Mikolov:2013arxiv,Smith:2017iclr} or document-aligned comparable data \cite{Vulic:2015sigir}.\footnote{For a complete overview we refer the reader to a recent survey \cite{Ruder:2017arxiv}.} 

Recently, methods for inducing shared cross-lingual embedding spaces without the need for any bilingual signal (not even word translation pairs) have been proposed \cite{Artetxe:2017acl,Conneau:2018iclr}. These methods exploit inherent structural similarities of induced monolingual embedding spaces to learn vector space transformations that align the source language space to the target language space, with strong results observed for bilingual lexicon extraction. In this work, we show that these unsupervised cross-lingual word embeddings offer strong support to the construction of fully unsupervised ad-hoc CLIR models. We propose two different CLIR models: \textbf{1)} term-by-term translation through the shared cross-lingual space, and \textbf{2)} query and document representations as IDF-weighted sums of constituent word vectors. To the best of our knowledge, our CLIR methodology is the first to allow the construction of CLIR models without any bilingual data and supervision at all, relying solely on monolingual corpora. 
%
Experimental evaluation on standard CLEF CLIR data for three different language pairs shows that the proposed fully unsupervised CLIR models outperform competitive baselines and models that exploit word translation pairs or comparable corpora. Our CLIR code and multilingual embedding spaces are publicly
available at: https://github.com/rlitschk/UnsupCLIR.

\label{s:intro}


\section{Methodology}
The proposed unsupervised CLIR models rely on the existence of a shared cross-lingual word embedding space in which all vocabulary terms of both languages are placed. We first outline three methods for the shared space induction, with a focus on the unsupervised method. We then explain in detail the query and document representations as well as the ranking functions of our CLIR models.        


\subsection{Cross-Lingual Word Vector Spaces}


For our proposed CLIR models, we investigate cross-lingual embedding spaces produced with state-of-the-art representative methods requiring different amount and type of bilingual supervision: \textbf{1)} document-aligned comparable data \cite{Vulic:2015sigir}, \textbf{2)} word translation pairs \cite{Smith:2017iclr}; and \textbf{3)} \textit{no bilingual data at all} \cite{Conneau:2018iclr}.    

\paragraph{Cross-Lingual Embeddings from Comparable Documents (CL-CD)} 

The BWE Skip-Gram (BWESG) model from \citet{Vulic:2015sigir} exploits large document-aligned comparable corpora (e.g., Wikipedia). BWESG first creates a merged corpus of bilingual pseudo-documents by intertwining pairs of available comparable documents. Then it applies a standard monolingual log-linear Skip-Gram model with negative sampling (SGNS) \cite{Mikolov:2013nips} on the merged corpus in which words have bilingual contexts instead of monolingual ones. 


\paragraph{Cross-Lingual Embeddings from Word Translation Pairs (CL-WT)} 

This class of models \cite{Mikolov:2013arxiv,Smith:2017iclr,Artetxe:2017acl} focuses on learning the projections (i.e., mappings) between independently trained monolingual embedding spaces. Let $\{v^S_{w^i}\}_{i = 1}^{V_S}, v^S_{w^i} \in \mathbb{R}^{\mathit{ds}}$ be the monolingual word embedding space of the source language $L_S$ with $V_S$ vectors, and $\{v^T_{w^i}\}_{i = 1}^{V_T}, v^T_{w^i} \in \mathbb{R}^{\mathit{dt}}$ the monolingual space for the target language $L_T$ containing $V_T$ vectors; $\mathit{ds}$ and $\mathit{dt}$ are the respective space dimensionalities. The models learn a parametrized mapping function $f(v | \theta)$ that projects the source language vectors into the target space: $f(v | \theta): \mathbb{R}^{\mathit{ds}} \rightarrow \mathbb{R}^{\mathit{dt}}.$ The projection parameters $\theta$ are learned using the training set of $K$ word translation pairs: $\{w^S_i, w^T_i\}_{i = 1}^K$, typically via second-order stochastic optimisation techniques.

According to the comparative evaluation from \cite{Ruder:2017arxiv}, all projection-based methods for inducing cross-lingual embedding spaces perform similarly. We therefore opt for the recent model of \citet{Smith:2017iclr} to serve as a baseline, due to its competitive performance, large coverage, and readily available implementation.\footnote{https://github.com/Babylonpartners/fastText\_multilingual} Technically, the method of \citet{Smith:2017iclr} learns two projection functions $f_S(v_S | \theta_S)$ and $f_S(v_T | \theta_T)$, projecting the source and target monolingual embedding spaces, respectively, to the new shared space.

%

\paragraph{Cross-Lingual Embeddings without Bilingual Supervision (CL-UNSUP)}

Most recently, \citet{Conneau:2018iclr} have proposed an adversarial learning-based model in order to automatically, in a fully unsupervised fashion, create word translation pairs that can then be used to learn the same projection functions $f_S$ and $f_T$ as in the model of \citet{Smith:2017iclr}. Let $X$ be the set of all monolingual word embeddings from the source language, and $Y$ the set of all target language embeddings. In the first, adversarial learning step, they jointly learn (1) the projection matrix $W$ that maps one embedding space to the other and (2) the parameters of the discriminator model which, given an embedding vector (either $Wx$ where $x \in X$, or $y \in Y$) needs to predict whether it is an original vector from the target embedding space ($y$),nor a vector from the source embedding space mapped via projection $W$ to the target embedding space ($Wx$). The discriminator model is a multi-layer perceptron network. In the second step, the projection matrix $W$ trained with adversarial objective is used to find the mutual nearest neighbors between the two vocabularies -- this set of automatically obtained word translation pairs becomes a synthetic training set for the refined projection functions $f_S$ and $f_T$ computed via the SVD-based method similar to the previously described model of \citet{Smith:2017iclr}.                          

\subsection{Unsupervised CLIR Models}

With the induced cross-lingual spaces we can directly measure semantic similarity of words from the two languages, but we still need to define how to represent queries and documents. To this end, we outline two models that exploit the induced cross-lingual embedding spaces for CLIR tasks.   

\paragraph{BWE aggregation model (BWE-AGG)} In the first approach, we derive the cross-lingual embeddings of queries and documents by aggregating the cross-lingual embeddings of their constituent terms. Let $\overrightarrow{t}$ be the embedding of the term $t$, obtained from the cross-lingual embedding space and let d = $\{t_1, t_2, \dots, t_{N_d}\}$ be a document from the collection consisting of $N_d$ terms. The embedding of the document $d$ in the shared space can then be computed as: 
\begin{equation*}
\overrightarrow{d} = \overrightarrow{t1} \circ \overrightarrow{t2} \circ \ldots \circ \overrightarrow{t_{N_d}} 
\end{equation*}
\noindent where $\circ$ is a semantic composition operator: it aggregates constituent term embeddings into a document embedding.\footnote{There is a large number of options for the composition operator, ranging from unsupervised operations like addition and element-wise multiplication \cite{mitchell2008vector} to complex parametrized (e.g., tensor-based) composition functions \cite{milajevs-EtAl:2014:EMNLP2014}. We discard the parametrized composition functions because they require parameter optimization through supervision, and we are interested in \textit{fully unsupervised resource-lean CLIR}.} We opt for vector addition as composition for two reasons: 1) word embedding spaces exhibit linear linguistic regularities \cite{Mikolov:2013naacl}, and 2) addition displays robust performance in compositional and IR tasks \cite{mitchell2008vector,Vulic:2015sigir}. A representation of the query vector $\overrightarrow{q}$ is then the sum of  embeddings of constituent terms: $\overrightarrow{q} = \sum_{i = 1}^{N_q}{\overrightarrow{t^q_i}}$. To obtain document representations, we compare two aggregation functions. First, we experiment with a simple non-weighted addition (\textit{BWE-Agg-Add}): $\overrightarrow{d} = \sum_{i = 1}^{N_d}{\overrightarrow{t^d_i}}$. Second, we use weighted addition where each term's embedding is weighted with the term's inverse document frequency (IDF) (\textit{BWE-Agg-IDF}): $\overrightarrow{d} = \sum_{i = 1}^{N_d}{\mathit{idf}(t^d_i) \cdot \overrightarrow{t^d_i}}$. BWE-Agg-IDF relies on the common assumption that not all terms equally contribute to the document meaning: it emphasizes vectors of more document-specific terms.\footnote{Note that with both variants of BWE-Agg, we effectively ignore both query and document terms that are not represented in the cross-lingual embedding space. } Finally, we compute the relevance score simply as the cosine similarity between query and document embeddings in the shared cross-lingual space:
$\mathit{rel}_{\mathit{Agg}}(q, d) = \frac{\overrightarrow{q}\cdot\overrightarrow{d}}{\lVert\overrightarrow{q}\rVert\cdot\lVert\overrightarrow{d}\rVert}.$


%
%
%

%
%
%
%
  
%

\paragraph{Term-by-term query translation model (TbT-QT)}

Our second CLIR model exploits the cross-lingual word embedding space in a different manner: it performs a term-by-term translation of the query into the language of the document collection relying solely on the shared cross-lingual space. Each source language query term $t^q$ is replaced by the target language term $\mathit{tr}(t^q)$, that is, its cross-lingual nearest neighbour in the embedding space. The cosine similarity is used for computing cross-lingual semantic similarities of terms. In other words, the query $q = \{t^q_1, t^q_2, \dots, t^q_{N_q}\}$ in $L_S$ is substituted by the query $q' = \{\mathit{tr}(t^q_1), \mathit{tr}(t^q_2), \dots, \mathit{tr}(t^q_{N_q})\}$ in $L_T$.\footnote{If the representation of a query term  $t^q_i$ is not present in the cross-lingual embedding space, we retain the query term $t^q_i$ itself. We have also attempted eliminating out-of-vocabulary query terms, but the former consistently leads to better performance.}

%
%

By effectively transforming a CLIR task into a monolingual IR task, we can apply any of the traditional IR ranking functions designed for sparse text representations. We opt for the ubiquitous query likelihood model \cite{ponte1998language}, smoothing the unigram language model of individual documents with the unigram language model of the entire collection, using the Dirichlet smoothing scheme \cite{zhai2004study}:  
\begin{equation*}
\textstyle{\mathit{rel}_{\mathit{TbT}}(q', d) = \prod_{i = 1}^{N_{q'}}{\lambda \cdot P(t^{q'}_i|d) + (1 - \lambda) \cdot P(t^{q'}_i|D)}}.
\end{equation*}
\noindent $P(t^{q'}_i|d)$ is the maximum likelihood estimate (MLE) of $t^{q'}_i$ probability based on the document $d$, $P(t^{q'}_i|D)$ is the MLE of term's probability based on the target collection $D$, and $\lambda = N_d / (N_d + \mu)$ determines the ratio between the contributions of the local and global language model, with $N_d$ being the document length and $\mu$ the parameter of Dirichlet smoothing ($=1000$ \cite{zhai2004study}). Note that the \textit{TbT-QT} model with unsupervised cross-lingual word embeddings is again a fully unsupervised CLIR framework.
\label{s:methodology}

\section{Experimental Setup}
\paragraph{Language Pairs and Training Data}
We experiment with three language pairs of varying degree of similarity: English (EN) -- \{Dutch (NL), Italian (IT), Finnish (FI)\}.\footnote{English and Dutch are Germanic languages, Italian is a Romance language, whereas Finnish is an Uralic language (i.e., not Indo-European)} We use precomputed monolingual \textsc{fastText} vectors \cite{Bojanowski:2017tacl} (available online)\footnote{https://github.com/facebookresearch/fastText} as monolingual word embeddings required by CL-WT and CL-UNSUP embedding models. For the CL-CD embeddings, the BWESG model trains on full document-aligned Wikipedias\footnote{http://linguatools.org/tools/corpora/wikipedia-comparable-corpora/} using SGNS with suggested parameters from prior work \cite{Vulic:2016jair}: $15$ negative samples, global decreasing learning rate is $.025$, subsampling rate is $1e-4$, window size is 16.


The CL-WT embeddings of \citet{Smith:2017iclr} use 10K translation pairs obtained from Google Translate to learn the linear mapping functions. The CL-UNSUP training setup closely follows the default setup of \citet{Conneau:2018iclr}: we refer the reader to the original paper and the model implementation accessible online for more information and technical details.\footnote{https://github.com/facebookresearch/MUSE}

\paragraph{Test Collections and Queries}
We evaluate the models on the standard test collections from the CLEF 2000-2003 ad-hoc retrieval Test Suite.\footnote{http://catalog.elra.info/product\_info.php?products\_id=888} We select all NL, IT, and FI document collections from years 2001-2003\footnote{Finnish was included to CLEF evaluation only in 2002 and 2003.} and paired them with English queries from the respective year. The statistics for test collections are shown in Table~\ref{tbl:stats}.    
\setlength{\tabcolsep}{3.2pt}
\begin{table}[t]
\centering
\def\arraystretch{0.75}
\vspace{-0.0em}
{\small
\begin{tabularx}{\linewidth}{c ccc ccc ccc}
\toprule
& \multicolumn{3}{c}{2001} & \multicolumn{3}{c}{2002} & \multicolumn{3}{c}{2003} \\ 
\cmidrule(lr){2-4} \cmidrule(lr){5-7} \cmidrule(lr){8-10}
Lang. & \#doc & \#tok & \#rel & \#doc & \#tok & \#rel & \#doc & \#tok & \#rel \\ \midrule
NL & 190K & 29.6M & 24.5 & 190K & 29.6M & 37.2 & 190K & 29.6M & 28.2 \\
IT & 108K & 17.1M & 26.5 & 108K & 17.1M & 21.9 & 22.3M & 157K & 15.9 \\
FI & -- & -- & -- & 55K & 9.3M & 16.7 & 55K & 9.25M & 10.7 \\
\bottomrule
\end{tabularx}}
\caption{Basic statistics of used CLEF test collections: number of documents (\#doc), number of tokens (\#tok), and average number of relevant documents per query (\#rel).}
\label{tbl:stats}
\vspace{-4.5mm}
\end{table}
\noindent Following a standard practice \cite{Lavrenko:2002sigir,Vulic:2015sigir}, queries were created by concatenating the \textit{title} and the \textit{description} of each CLEF ``topic''. The test collections for years 2001-2003 respectively contain 50, 50, and 60 EN queries. Queries and documents were lowercased; stop words, punctuations and one-character words were removed.

\paragraph{Models in Comparison.}

We evaluate six different CLIR models, obtained by combining each of the three models for inducing cross-lingual word vector spaces -- \textit{CL-CD}, \textit{CL-WT}, and \textit{CL-UNSUP} -- with each of the two ranking models -- \textit{BWE-Agg} and \textit{TbT-QT}. For each cross-lingual vector space, we also evaluate an ensemble ranker that combines the two ranking functions: \textit{BWE-Agg-IDF} and \textit{TbT-QT}. If $r_1$ is the rank of document $d$ for query $q$ according to the \textit{TbT-QT} model and $r_2$ is the rank produced by \textit{BWE-Agg-IDF}, the ensemble ranker ranks the documents in the increasing order of the scores $\lambda \cdot r_1 + (1 - \lambda)\cdot r_2$. We evaluate ensembles with values $\lambda = 0.5$, i.e., with equal contributions of both models; and $\lambda = 0.7$, i.e., with more weight allocated to the more powerful \textit{TbT-QT} model (cf.~Table \ref{tbl:baseresults}). 
%
Additionally, we evaluate the standard query likelihood model (\textit{LM-UNI}) \cite{ponte1998language} with Dirichlet smoothing \cite{zhai2004study} as a direct baseline.\footnote{LM-UNI uses the same ranking function as TbT-QT, but without the prior term-by-term query translation via the cross-lingual embedding space. LM-UNI is more suitable for monolingual IR than for CLIR due to limited lexical overlap between languages.}
\label{s:experimental}

\section{Results and Discussion}
We show performance of all models in comparison on all test collections, reported in terms of the standard \textit{mean average precision} (MAP) measure in Table~\ref{tbl:baseresults}.     
\setlength{\tabcolsep}{13pt}
\begin{table*}[t]
\centering
\def\arraystretch{0.7}
\vspace{-0.0em}
{\normalsize
\begin{tabularx}{\linewidth}{l l cc ccc ccc}
\toprule
 &  & \multicolumn{3}{c}{EN$\rightarrow$NL} & \multicolumn{3}{c}{EN$\rightarrow$IT} & \multicolumn{2}{c}{EN$\rightarrow$FI} \\ 
\cmidrule(lr){3-5} \cmidrule(lr){6-8} \cmidrule(lr){9-10}
CL Embs & Model & 2001 & 2002 & 2003 & 2001 & 2002 & 2003 & 2002 & 2003 \\ \midrule
-- & LM-UNI & .119 & .196 & .136 & .085 & .167 & .137 & .111 & .142  \\ \midrule
 & BWE-Agg-Add & .111 & .138 & .137 & .087 & .114 & .147 & .026 & .084 \\ 
 & BWE-Agg-IDF & .144 & .203 & .189 & .127 & .157 & .188 & .082 & .125 \\
CL-CD & TbT-QT & .125 & .196 & .120 & .106 & .148 & .143 & \textbf{.176} & .140 \\ \cdashline{2-10}
& Ensemble ($\lambda = 0.5$) & .145 & .216 & .174 & .120 & .183 & .216 & .179 & .189 \\ 
& Ensemble ($\lambda = 0.7$) & .142 & .216 & .180 & .127 & .180 & .207 & .183 & .197 \\ \midrule
 & BWE-Agg-Add & .149 & .168 & .203 & .138 & .155 & .236 & .078 & .217 \\
 & BWE-Agg-IDF & .185 & .196 & .243 & .169 & .166 & .248 & .086 & .204 \\
CL-WT & TbT-QT & .159 & .164 & .176 & .129 & .150 & .218 & .095 & .095 \\ \cdashline{2-10} 
 & Ensemble ($\lambda = 0.5$) & .202 & .198 & .280 & .187 & .168 & .228 & .117 & .190  \\ 
& Ensemble ($\lambda = 0.7$) & .202 & .198 & .263 & .181 & .171 & .230 & .120 & .164 \\ \midrule
 & BWE-Agg-Add & .125 & .153 & .198 & .119 & .126 & .213 & .078 & .239 \\
 & BWE-Agg-IDF & .172 & .204 & .250 & .157 & .161 & .253 & .102 & .223 \\
CL-UNSUP & TbT-QT & \textbf{.229} & \textbf{.257} & \textbf{.299} & \textbf{.232} & \textbf{.257} & \textbf{.345} & .145 & \textbf{.243} \\ \cdashline{2-10} 
& Ensemble ($\lambda = 0.5$) & .258 & .300 & .330 & .225 & .248 & .325 & .154 & \textbf{.307} \\ 
& Ensemble ($\lambda = 0.7$) & \textbf{.259} & \textbf{.303} & \textbf{.336} & \textbf{.236} & .253 & \textbf{.347} & .151 & \textbf{.307} \\
\bottomrule
\end{tabularx}}
\caption{CLIR performance on all three test language pairs for all models in comparison (MAP scores reported).}
\label{tbl:baseresults}
\vspace{-4.5mm}
\end{table*}

\paragraph{Unsupervised vs. Supervised CLIR}

First, CLIR models based on CL-WT embeddings (the bilingual signal are word translation pairs) outperform models based on CL-CD (requiring document-aligned data) on average. This is an encouraging finding, as word translations pairs are easier to obtain than document-aligned comparable corpora. Most importantly, the unsupervised CL-UNSUP+TbT-QT CLIR model displays peak performance on all but one test collection (EN-FI, 2002). We find this to be a very important result: it shows that we can perform robust CLIR without any cross-lingual information, that is, by relying purely on monolingual data. 


\paragraph{Ensemble CLIR Models}

Ensembles generally outperform the best-performing individual CLIR models, and for some test collections (e.g., EN$\rightarrow$NL 2002, EN$\rightarrow$FI 2003) by a wide margin. For the \textit{CL-CD} and \textit{CL-WT} spaces, we observe similar results for both values of the interpolation factor ($\lambda = 0.5$ and $\lambda = 0.7$). This is not surprising, since the single models \textit{BWE-Agg-IDF} and \textit{TbT-QT} exhibit similar performance for \textit{CL-CD} and \textit{CL-WT}. In contrast, the combined model with $\lambda = 0.7$ (i.e., more weight for the \textit{TbT-QT} ranking) yields larger performance gains for CL-UNSUP spaces, for which the \textit{TbT-QT} model consistently outperforms \textit{BWE-Agg-IDF}.     

\paragraph{Language Similarity and Aggregation}

The results in Table \ref{tbl:baseresults} imply that the proximity of CLIR languages plays a role only to a certain extent. Most models do exhibit lower performance for EN$\rightarrow$FI than for the other two language pairs: this is expected since Finnish is lexically and typologically more distant from English than Italian and Dutch. However, even though NL is linguistically closer to EN than IT, for the unsupervised CLIR models we generally observe slightly better performance for EN$\rightarrow$IT than for EN$\rightarrow$NL. We speculate that this is due to the compounding phenomenon in word formation, which is present in NL, but is not a property of EN and IT. The reported performance on bilingual lexicon extraction (BLE) using cross-lingual embedding spaces is also lower for EN-NL compared to EN-IT (see, e.g., \cite{Smith:2017iclr}). We observe the same pattern (4-5\% lower BLE performance for EN-NL than for EN-IT) with the CL-UNSUP embedding spaces.

The weighted variant of BWE-Agg (BWE-Agg-IDF) outperforms the simpler non-weighted summation model (BWE-Agg-Add) across the board. These results suggest that the common IR assumption about document-specific terms being more important than the terms occurring collection-wide is also valid for constructing dense document representations by summing word embeddings. 


       
\label{s:results}

\section{Conclusion}
We have presented a fully unsupervised CLIR framework that leverages unsupervised cross-lingual word embeddings induced solely on the basis of monolingual corpora. We have shown the ability of our models to retrieve relevant content cross-lingually without any bilingual data at all, by reporting competitive performance on standard CLEF CLIR evaluation data for three test language pairs. This unsupervised framework holds promise to support and guide the development of effective CLIR models for language pairs and domains where parallel data are scarce or unavailable.
\label{s:conclusion}

\bibliographystyle{ACM-Reference-Format}
\bibliography{sigir2018_refs}


\begin{thebibliography}{23}


\ifx \showCODEN    \undefined \def \showCODEN     #1{\unskip}     \fi
\ifx \showDOI      \undefined \def \showDOI       #1{#1}\fi
\ifx \showISBNx    \undefined \def \showISBNx     #1{\unskip}     \fi
\ifx \showISBNxiii \undefined \def \showISBNxiii  #1{\unskip}     \fi
\ifx \showISSN     \undefined \def \showISSN      #1{\unskip}     \fi
\ifx \showLCCN     \undefined \def \showLCCN      #1{\unskip}     \fi
\ifx \shownote     \undefined \def \shownote      #1{#1}          \fi
\ifx \showarticletitle \undefined \def \showarticletitle #1{#1}   \fi
\ifx \showURL      \undefined \def \showURL       {\relax}        \fi
\providecommand\bibfield[2]{#2}
\providecommand\bibinfo[2]{#2}
\providecommand\natexlab[1]{#1}
\providecommand\showeprint[2][]{arXiv:#2}

\bibitem[\protect\citeauthoryear{Artetxe, Labaka, and Agirre}{Artetxe
  et~al\mbox{.}}{2017}]%
        {Artetxe:2017acl}
\bibfield{author}{\bibinfo{person}{Mikel Artetxe}, \bibinfo{person}{Gorka
  Labaka}, {and} \bibinfo{person}{Eneko Agirre}.}
  \bibinfo{year}{2017}\natexlab{}.
\newblock \showarticletitle{Learning bilingual word embeddings with (almost) no
  bilingual data}. In \bibinfo{booktitle}{\emph{ACL}}.
  \bibinfo{pages}{451--462}.
\newblock


\bibitem[\protect\citeauthoryear{Bojanowski and et~al}{Bojanowski and
  et~al}{2017}]%
        {Bojanowski:2017tacl}
\bibfield{author}{\bibinfo{person}{Piotr Bojanowski} {and}
  \bibinfo{person}{Edouard~Grave et al}.} \bibinfo{year}{2017}\natexlab{}.
\newblock \showarticletitle{Enriching Word Vectors with Subword Information}.
\newblock \bibinfo{journal}{\emph{Transactions of the ACL}}
  \bibinfo{volume}{5} (\bibinfo{year}{2017}), \bibinfo{pages}{135--146}.
\newblock


\bibitem[\protect\citeauthoryear{Conneau, Lample, Ranzato, Denoyer, and
  J{\'{e}}gou}{Conneau et~al\mbox{.}}{2018}]%
        {Conneau:2018iclr}
\bibfield{author}{\bibinfo{person}{Alexis Conneau}, \bibinfo{person}{Guillaume
  Lample}, \bibinfo{person}{Marc'Aurelio Ranzato}, \bibinfo{person}{Ludovic
  Denoyer}, {and} \bibinfo{person}{Herv{\'{e}} J{\'{e}}gou}.}
  \bibinfo{year}{2018}\natexlab{}.
\newblock \showarticletitle{Word Translation Without Parallel Data}. In
  \bibinfo{booktitle}{\emph{ICLR}}.
\newblock


\bibitem[\protect\citeauthoryear{Franco-Salvador, Rosso, and
  Navigli}{Franco-Salvador et~al\mbox{.}}{2014}]%
        {franco2014knowledge}
\bibfield{author}{\bibinfo{person}{Marc Franco-Salvador},
  \bibinfo{person}{Paolo Rosso}, {and} \bibinfo{person}{Roberto Navigli}.}
  \bibinfo{year}{2014}\natexlab{}.
\newblock \showarticletitle{A knowledge-based representation for cross-language
  document retrieval and categorization}. In \bibinfo{booktitle}{\emph{EACL}}.
  \bibinfo{pages}{414--423}.
\newblock


\bibitem[\protect\citeauthoryear{Hermann and Blunsom}{Hermann and
  Blunsom}{2014}]%
        {hermann-blunsom:2014:P14-1}
\bibfield{author}{\bibinfo{person}{Karl~Moritz Hermann} {and}
  \bibinfo{person}{Phil Blunsom}.} \bibinfo{year}{2014}\natexlab{}.
\newblock \showarticletitle{Multilingual Models for Compositional Distributed
  Semantics}. In \bibinfo{booktitle}{\emph{ACL}}. \bibinfo{pages}{58--68}.
\newblock


\bibitem[\protect\citeauthoryear{Landauer and Dumais}{Landauer and
  Dumais}{1997}]%
        {Landauer:19971}
\bibfield{author}{\bibinfo{person}{Thomas~K. Landauer} {and}
  \bibinfo{person}{Susan~T. Dumais}.} \bibinfo{year}{1997}\natexlab{}.
\newblock \showarticletitle{Solutions to {P}lato's problem: {T}he {L}atent
  {S}emantic {A}nalysis Theory of Acquisition, Induction, and Representation of
  Knowledge}.
\newblock \bibinfo{journal}{\emph{Psychological Review}} \bibinfo{volume}{104},
  \bibinfo{number}{2} (\bibinfo{year}{1997}), \bibinfo{pages}{211--240}.
\newblock


\bibitem[\protect\citeauthoryear{Lavrenko, Choquette, and Croft}{Lavrenko
  et~al\mbox{.}}{2002}]%
        {Lavrenko:2002sigir}
\bibfield{author}{\bibinfo{person}{Victor Lavrenko}, \bibinfo{person}{Martin
  Choquette}, {and} \bibinfo{person}{W.~Bruce Croft}.}
  \bibinfo{year}{2002}\natexlab{}.
\newblock \showarticletitle{Cross-lingual relevance models}. In
  \bibinfo{booktitle}{\emph{SIGIR}}. \bibinfo{pages}{175--182}.
\newblock


\bibitem[\protect\citeauthoryear{Levow, Oard, and Resnik}{Levow
  et~al\mbox{.}}{2005}]%
        {Levow:2005ipm}
\bibfield{author}{\bibinfo{person}{Gina-Anne Levow},
  \bibinfo{person}{Douglas~W. Oard}, {and} \bibinfo{person}{Philip Resnik}.}
  \bibinfo{year}{2005}\natexlab{}.
\newblock \showarticletitle{Dictionary-Based Techniques for Cross-Lingual IR}.
\newblock \bibinfo{journal}{\emph{IP \& M}} \bibinfo{volume}{41},
  \bibinfo{number}{3} (\bibinfo{year}{2005}), \bibinfo{pages}{523--547}.
\newblock


\bibitem[\protect\citeauthoryear{Martino and et~al.}{Martino and
  et~al.}{2017}]%
        {Martino:2017sigir}
\bibfield{author}{\bibinfo{person}{Giovanni Da~San Martino} {and}
  \bibinfo{person}{Salvatore~Romeo et al.}} \bibinfo{year}{2017}\natexlab{}.
\newblock \showarticletitle{Cross-language question re-ranking}. In
  \bibinfo{booktitle}{\emph{SIGIR}}. \bibinfo{pages}{1145--1148}.
\newblock


\bibitem[\protect\citeauthoryear{Mikolov and et~al}{Mikolov and et~al}{2013}]%
        {Mikolov:2013nips}
\bibfield{author}{\bibinfo{person}{Tomas Mikolov} {and}
  \bibinfo{person}{Ilya~Sutskever et al}.} \bibinfo{year}{2013}\natexlab{}.
\newblock \showarticletitle{Distributed Representations of Words and Phrases
  and their Compositionality}. In \bibinfo{booktitle}{\emph{NIPS}}.
  \bibinfo{pages}{3111--3119}.
\newblock


\bibitem[\protect\citeauthoryear{Mikolov, Le, and Sutskever}{Mikolov
  et~al\mbox{.}}{2013a}]%
        {Mikolov:2013arxiv}
\bibfield{author}{\bibinfo{person}{Tomas Mikolov}, \bibinfo{person}{Quoc~V.
  Le}, {and} \bibinfo{person}{Ilya Sutskever}.}
  \bibinfo{year}{2013}\natexlab{a}.
\newblock \showarticletitle{Exploiting Similarities among Languages for Machine
  Translation}.
\newblock \bibinfo{journal}{\emph{CoRR}}  \bibinfo{volume}{abs/1309.4168}
  (\bibinfo{year}{2013}).
\newblock


\bibitem[\protect\citeauthoryear{Mikolov, Yih, and Zweig}{Mikolov
  et~al\mbox{.}}{2013b}]%
        {Mikolov:2013naacl}
\bibfield{author}{\bibinfo{person}{Tomas Mikolov}, \bibinfo{person}{Wen-tau
  Yih}, {and} \bibinfo{person}{Geoffrey Zweig}.}
  \bibinfo{year}{2013}\natexlab{b}.
\newblock \showarticletitle{Linguistic Regularities in Continuous Space Word
  Representations}. In \bibinfo{booktitle}{\emph{NAACL-HLT}}.
  \bibinfo{pages}{746--751}.
\newblock


\bibitem[\protect\citeauthoryear{Milajevs and et~al}{Milajevs and
  et~al}{2014}]%
        {milajevs-EtAl:2014:EMNLP2014}
\bibfield{author}{\bibinfo{person}{Dmitrijs Milajevs} {and}
  \bibinfo{person}{Dimitri~Kartsaklis et al}.} \bibinfo{year}{2014}\natexlab{}.
\newblock \showarticletitle{Evaluating Neural Word Representations in
  Tensor-Based Compositional Settings}. In \bibinfo{booktitle}{\emph{EMNLP}}.
  \bibinfo{pages}{708--719}.
\newblock


\bibitem[\protect\citeauthoryear{Mitchell and Lapata}{Mitchell and
  Lapata}{2008}]%
        {mitchell2008vector}
\bibfield{author}{\bibinfo{person}{Jeff Mitchell} {and}
  \bibinfo{person}{Mirella Lapata}.} \bibinfo{year}{2008}\natexlab{}.
\newblock \showarticletitle{Vector-based models of semantic composition}. In
  \bibinfo{booktitle}{\emph{ACL-HLT}}. \bibinfo{pages}{236--244}.
\newblock


\bibitem[\protect\citeauthoryear{Mitra and Craswell}{Mitra and
  Craswell}{2017}]%
        {Mitra:2017arxiv}
\bibfield{author}{\bibinfo{person}{Bhaskar Mitra} {and} \bibinfo{person}{Nick
  Craswell}.} \bibinfo{year}{2017}\natexlab{}.
\newblock \showarticletitle{Neural Models for Information Retrieval}.
\newblock \bibinfo{journal}{\emph{CoRR}}  \bibinfo{volume}{abs/1705.01509}
  (\bibinfo{year}{2017}).
\newblock


\bibitem[\protect\citeauthoryear{Nie}{Nie}{2010}]%
        {Nie:2010book}
\bibfield{author}{\bibinfo{person}{Jian-Yun Nie}.}
  \bibinfo{year}{2010}\natexlab{}.
\newblock \bibinfo{booktitle}{\emph{Cross-Language Information Retrieval}}.
\newblock


\bibitem[\protect\citeauthoryear{Ponte and Croft}{Ponte and Croft}{1998}]%
        {ponte1998language}
\bibfield{author}{\bibinfo{person}{Jay~M. Ponte} {and}
  \bibinfo{person}{W.~Bruce Croft}.} \bibinfo{year}{1998}\natexlab{}.
\newblock \showarticletitle{A language modeling approach to information
  retrieval}. In \bibinfo{booktitle}{\emph{SIGIR}}. ACM,
  \bibinfo{pages}{275--281}.
\newblock


\bibitem[\protect\citeauthoryear{Ruder, Vuli\'{c}, and S{\o}gaard}{Ruder
  et~al\mbox{.}}{2017}]%
        {Ruder:2017arxiv}
\bibfield{author}{\bibinfo{person}{Sebastian Ruder}, \bibinfo{person}{Ivan
  Vuli\'{c}}, {and} \bibinfo{person}{Anders S{\o}gaard}.}
  \bibinfo{year}{2017}\natexlab{}.
\newblock \showarticletitle{A Survey of Cross-Lingual Embedding Models}.
\newblock \bibinfo{journal}{\emph{CoRR}}  \bibinfo{volume}{abs/1706.04902}
  (\bibinfo{year}{2017}).
\newblock


\bibitem[\protect\citeauthoryear{Smith, Turban, Hamblin, and Hammerla}{Smith
  et~al\mbox{.}}{2017}]%
        {Smith:2017iclr}
\bibfield{author}{\bibinfo{person}{Samuel~L. Smith},
  \bibinfo{person}{David~H.P. Turban}, \bibinfo{person}{Steven Hamblin}, {and}
  \bibinfo{person}{Nils~Y. Hammerla}.} \bibinfo{year}{2017}\natexlab{}.
\newblock \showarticletitle{Offline Bilingual Word Vectors, Orthogonal
  Transformations and the Inverted Softmax}. In
  \bibinfo{booktitle}{\emph{ICLR}}.
\newblock


\bibitem[\protect\citeauthoryear{Sorg and Cimiano}{Sorg and Cimiano}{2012}]%
        {sorg2012exploiting}
\bibfield{author}{\bibinfo{person}{Philipp Sorg} {and} \bibinfo{person}{Philipp
  Cimiano}.} \bibinfo{year}{2012}\natexlab{}.
\newblock \showarticletitle{Exploiting Wikipedia for cross-lingual and
  multilingual information retrieval}.
\newblock \bibinfo{journal}{\emph{DKE}}  \bibinfo{volume}{74}
  (\bibinfo{year}{2012}), \bibinfo{pages}{26--45}.
\newblock


\bibitem[\protect\citeauthoryear{Vuli\'{c} and Moens}{Vuli\'{c} and
  Moens}{2015}]%
        {Vulic:2015sigir}
\bibfield{author}{\bibinfo{person}{Ivan Vuli\'{c}} {and} \bibinfo{person}{Sien
  Moens}.} \bibinfo{year}{2015}\natexlab{}.
\newblock \showarticletitle{Monolingual and Cross-lingual Information Retrieval
  Models Based on (Bilingual) Word Embeddings}. In
  \bibinfo{booktitle}{\emph{SIGIR}}. \bibinfo{pages}{363--372}.
\newblock


\bibitem[\protect\citeauthoryear{Vuli\'{c} and Moens}{Vuli\'{c} and
  Moens}{2016}]%
        {Vulic:2016jair}
\bibfield{author}{\bibinfo{person}{Ivan Vuli\'{c}} {and} \bibinfo{person}{Sien
  Moens}.} \bibinfo{year}{2016}\natexlab{}.
\newblock \showarticletitle{Bilingual Distributed Word Representations from
  Document-Aligned Comparable Data}.
\newblock \bibinfo{journal}{\emph{JAIR}}  \bibinfo{volume}{55}
  (\bibinfo{year}{2016}), \bibinfo{pages}{953--994}.
\newblock


\bibitem[\protect\citeauthoryear{Zhai and Lafferty}{Zhai and Lafferty}{2004}]%
        {zhai2004study}
\bibfield{author}{\bibinfo{person}{Chengxiang Zhai} {and} \bibinfo{person}{John
  Lafferty}.} \bibinfo{year}{2004}\natexlab{}.
\newblock \showarticletitle{A study of smoothing methods for language models
  applied to information retrieval}.
\newblock \bibinfo{journal}{\emph{ACM Transactions on Information Systems}}
  \bibinfo{volume}{22}, \bibinfo{number}{2} (\bibinfo{year}{2004}),
  \bibinfo{pages}{179--214}.
\newblock


\end{thebibliography}

\end{document}